\pdfoutput=1

\documentclass[letterpaper, 10 pt, conference]{ieeeconf}
\IEEEoverridecommandlockouts
\usepackage{cite}
\usepackage{amsmath,amssymb,amsfonts}
\usepackage{dsfont}
\usepackage{mathtools}
\usepackage{graphicx}
\usepackage{textcomp}
\usepackage{xcolor}
\usepackage{tikz}
\usepackage{pgfplots}
\pgfplotsset{compat=newest}
\usetikzlibrary{spy, arrows, arrows.meta, positioning, shapes, shapes.geometric, external, shadows, babel, calc, backgrounds, intersections, angles}
\usepackage{booktabs}
\usepackage{multirow}
\usepackage{makecell}
\usepackage{diagbox}
\usepackage{siunitx}
\usepackage{subcaption}

\makeatletter
\let\NAT@parse\undefined
\makeatother
\usepackage[hidelinks]{hyperref}

\definecolor{clutterPredictedClr}{HTML}{BF0016}
\definecolor{clutterMissedClr}{HTML}{FF3500}
\definecolor{objectPredictedClr}{HTML}{001EFF}
\definecolor{objectMissedClr}{HTML}{33B1FF}
\definecolor{stationaryPredictedClr}{HTML}{5A5A5A}
\definecolor{stationaryMissedClr}{HTML}{BBBBBB}
\definecolor{ignore1Clr}{HTML}{C1EAAF}
\definecolor{ignore2Clr}{HTML}{C1EAAF}

\pgfplotsset{
	colormap={cm_radar_dl_gt}{color(0)=(clutterPredictedClr), color(1)=(objectPredictedClr), color(2)=(stationaryPredictedClr), color(3)=(ignore1Clr)}
}

\pgfplotsset{
	colormap={cm_radar_dl_confusion}{color(0)=(clutterPredictedClr), color(1)=(clutterMissedClr), color(2)=(objectPredictedClr), color(3)=(objectMissedClr), color(4)=(stationaryPredictedClr), color(5)=(stationaryMissedClr), color(6)=(ignore1Clr), color(7)=(ignore2Clr)}
}

\makeatletter \newcommand{\pgfplotsdrawaxis}{\pgfplots@draw@axis} \makeatother

\pgfplotsset{axis line on top/.style={
		axis line style=transparent,
		axis on top=false,
		after end axis/.append code={
			\pgfplotsset{axis line style=opaque,
				ticklabel style=transparent,
				tick style=transparent,
				grid=none}
			\pgfplotsdrawaxis}
	}
}

\newcommand{\plotRadarDetectionListGT}[1] {
	\addplot
	[
	scatter,
	only marks,
	scatter src=explicit symbolic,
	scatter/classes={
		0={mark=*, clutterPredictedClr},
		1={mark=*, objectPredictedClr},
		2={mark=*, mark size=0.75pt, stationaryPredictedClr},
		3={mark=*, ignore1Clr}
	},
	mark size=1.0pt
	]
	table
	[
	x expr=\thisrow{y},
	y expr=\thisrow{x},
	meta=gt_label,
	col sep=comma
	] {#1};

	\addplot
	[
	point meta=explicit symbolic,
	quiver={
		u=\thisrow{v_y},
		v=\thisrow{v_x},
		scale arrows=0.6,
		colored
	},
	colormap access=direct,
	-{Stealth[length=3.0pt, width=2.3pt]}
	]
	table
	[
	x expr=\thisrow{y},
	y expr=\thisrow{x},
	meta=gt_label,
	col sep=comma,
	restrict expr to domain={pow(\thisrow{v_x}, 2) + pow(\thisrow{v_y}, 2)}{8.0:+inf}
	] {#1};
}

\newcommand{\plotRadarDetectionListConfusion}[1] {
	\addplot
	[
	scatter,
	only marks,
	scatter src=explicit symbolic,
	scatter/classes={
		0={mark=*, clutterPredictedClr},
		1={mark=*, clutterMissedClr},
		2={mark=*, objectPredictedClr},
		3={mark=*, objectMissedClr},
		4={mark=*, mark size=0.6pt, stationaryPredictedClr},
		5={mark=*, mark size=0.6pt, stationaryMissedClr},
		6={mark=*, mark size=0.6pt, ignore1Clr, fill opacity=0.7, draw opacity=0},
		7={mark=*, mark size=0.6pt, ignore2Clr, fill opacity=0.7, draw opacity=0}
	},
	mark size=0.8pt
	]
	table
	[
	x expr=\thisrow{y},
	y expr=\thisrow{x},
	meta=confusion_value,
	col sep=comma
	] {#1};

	\addplot
	[
	point meta=explicit symbolic,
	quiver={
		u=\thisrow{v_y},
		v=\thisrow{v_x},
		scale arrows=0.6,
		every arrow/.append style={
			color=mapped color,
			opacity=(\pgfplotspointmeta < 6) * 1 + !(\pgfplotspointmeta < 6) * 0.7
		}
	},
	colormap access=direct,
	-{Stealth[length=2.6pt, width=2.0pt]}
	]
	table
	[
	x expr=\thisrow{y},
	y expr=\thisrow{x},
	meta=confusion_value,
	col sep=comma,
	restrict expr to domain={pow(\thisrow{v_x}, 2) + pow(\thisrow{v_y}, 2)}{8.0:+inf}
	] {#1};
}

\title{\LARGE \textbf{Tackling Clutter in Radar Data --\\Label Generation and Detection Using PointNet++}}

\author{Johannes Kopp\textsuperscript{1}, Dominik Kellner\textsuperscript{2}, Aldi Piroli\textsuperscript{1} and Klaus Dietmayer\textsuperscript{1}
\thanks{\textsuperscript{1}Institute of Measurement, Control and Microtechnology, Ulm University, Albert-Einstein-Allee 41, 89081 Ulm, Germany {\tt\small \{firstname\}.\{lastname\}@uni-ulm.de}}%
\thanks{\textsuperscript{2}BMW AG, Petuelring 130, 80809 Munich, Germany {\tt\small dominik.m.kellner@bmw.de}}%
}

\newcommand\copyrighttext{
	\footnotesize \textcopyright 2023 IEEE.  Personal use of this material is permitted.  Permission from IEEE must be obtained for all other uses, in any current or future media, including reprinting/republishing this material for advertising or promotional purposes, creating new collective works, for resale or redistribution to servers or lists, or reuse of any copyrighted component of this work in other works.}
\newcommand\copyrightnotice[1]{
	\tikzset{external/export next=false}
	\begin{tikzpicture}[remember picture,overlay]
		\node[anchor=north,yshift=-15pt] at (current page.north) {\parbox{\dimexpr\textwidth-1.0cm}{#1}};
	\end{tikzpicture}
	\vspace{-10pt}
}

\begin{document}

\maketitle
\copyrightnotice{\copyrighttext}
\thispagestyle{empty}
\pagestyle{empty}

\begin{abstract}

Radar sensors employed for environment perception, e.g. in autonomous vehicles, output a lot of unwanted clutter. These points, for which no corresponding real objects exist, are a major source of errors in following processing steps like object detection or tracking. We therefore present two novel neural network setups for identifying clutter. The input data, network architectures and training configuration are adjusted specifically for this task. Special attention is paid to the downsampling of point clouds composed of multiple sensor scans. In an extensive evaluation, the new setups display substantially better performance than existing approaches.
Because there is no suitable public data set in which clutter is annotated, we design a method to automatically generate the respective labels. By applying it to existing data with object annotations and releasing its code, we effectively create the first freely available radar clutter data set representing real-world driving scenarios. Code and instructions are accessible at \url{www.github.com/kopp-j/clutter-ds}.

\end{abstract}

\section{Introduction}

Radar sensors play an important role in the environment perception of modern autonomous robot systems. This is especially true for outdoor applications that require a large field of view of \SI{100}{m} or more. Here, radar stands out against other sensor technologies like camera and lidar due to its long range and robustness to adverse weather conditions such as rain or fog \cite{Engels2021}. Moreover, it allows for a direct measurement of the velocities of objects. For these reasons, research is driven particularly by the automotive sector, where radar sensors are vital to the development of new driver assistance systems and a continuously rising degree of automation.

\begin{figure}[!t]
	\centering
	\vspace*{0.20cm}
	\includegraphics{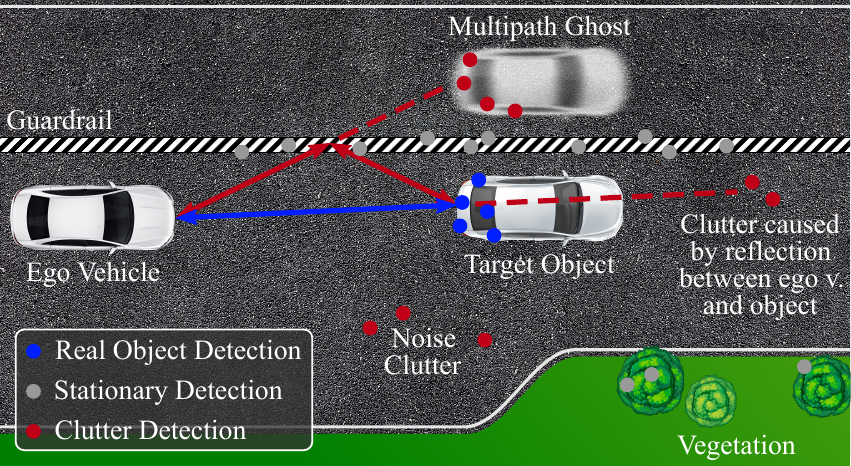}
	\caption{Visualization of exemplary causes of clutter. Arrows indicate propagation paths. Multipath ghosts are clutter resulting from specular reflection at objects like guardrails or concrete walls, which act like mirrors for radar waves.}
	\label{fig:overview_clutter}
\end{figure}

Automotive radar sensors perceive their surroundings by emitting electromagnetic waves and analyzing the reflections from obstacles in the propagation path. This way, the position of objects and their velocity in radial direction relative to the sensor are estimated. At the end of each ``scan'', a list of such detection points, also called targets, is output. What is problematic, however, is that this includes many detections that do not actually match any real object in the scene. This is what is called clutter\footnote[3]{Some authors refer to the same detections or a subset of them as ghosts, anomalies, artifacts or noise.}.
Clutter is not limited to (seemingly) random noise. Rather, a large part of it is linked to objects located elsewhere according to certain rules. Causes for this are e.g. indirect propagation paths or erroneous ambiguity resolution by the sensor's signal processing. Fig.~\ref{fig:overview_clutter} visualizes some of the corresponding effects.

Clutter linked to real objects behaves very similar to detections that correctly represent road users. This poses a substantial challenge to environment perception methods like object detection or tracking. Thus, it is desirable to identify and filter the problematic points previous to such steps. In this work, we present neural network setups realizing this functionality. Our main contributions are the following:
\begin{itemize}
	\item We describe a method for automatic label generation for an existing data set whose annotations do not yet incorporate clutter. The method's code is published.
	\item Building on a setup for the semantic segmentation of radar data, we design two novel network variants specifically tailored to the task of detecting clutter.
	\item We show how it is possible to generate a prediction for each relevant detection in a point cloud while keeping the number of processed points constant. To this end, we introduce accumulation-aware downsampling.
\end{itemize}

\section{Related Work} \label{section:related_work}

In recent years, clutter in automotive radar data has been recognized as an omnipresent and increasingly important issue \cite{Sun2020, Dickmann2016}. Causes for its occurrence and how they are reflected in a sensor's output are being studied e.g. \mbox{in \cite{Holder2019, Kopp2021, Scheiner2020, Kamann2018}.} \cite{Holder2019} describes and models a wide range of different effects and locates their origin in the processing chain. The most frequent and problematic propagation paths leading to clutter are analyzed in even more detail in~\cite{Kopp2021}. \cite{Scheiner2020} and \cite{Kamann2018} focus exclusively on specular reflection. Both contain a geometric model of the phenomenon. In addition to that, \cite{Scheiner2020} describes the effect on the level of electromagnetic signals, while the authors of \cite{Kamann2018} conduct practical experiments for verification.

Regarding the detection of clutter in automotive radar data, several different approaches exist. It must be noted, though, that many authors simplify the task by considering only a subset of possible clutter or restricting the variety of scenarios in their data.
For example, in both \cite{Roos2017} and \cite{Liu2020}, model-based algorithms are used to detect multipath ghosts only. The authors of \cite{Kraus2020} and \cite{Kraus2021} employ PointNet++ to distinguish between detections stemming from real objects, different types of multipath ghosts and other points (among further configurations). Their data is recorded with experimental radar sensors in a controlled environment. The ego vehicle is always stationary and, for the most part, there is only a single moving object. Other works limited to certain types of clutter are \cite{Griebel2021}, where PointNet++ is extended by a new grouping mechanism, and \cite{Wang2021}.
A wider scope is covered e.g. in \cite{Chamseddine2020}. Here, clutter of any kind is searched using an adapted PointNet architecture. The authors of \cite{Jin2021} compare a random forest, a convolutional neural network and PointNet++ regarding their effectiveness for clutter detection. But they restrict the evaluation to the area directly in front of the sensor. Lastly, we present a rule-based approach for identifying clutter in \cite{Kopp2021}.

The varying scopes and different definitions of erroneous detections in literature can mostly be explained by the fact that up until recently, there existed no public radar clutter data set. All authors used their own proprietary data. Even now, the only available data set~\cite{Kraus2021} is limited to multipath ghosts and does not adequately represent road traffic, as outlined above. These problems are solved by the label generation method we present in the next section. Comparable approaches have previously been realized in \cite{Chamseddine2020} and \cite{Wang2021}. However, both of them suffer from major drawbacks. The former requires a 3D radar sensor, something still rather uncommon, and the availability of additional lidar data for label generation. The second approach relies not only on lidar points but also existing annotations regarding the positions and dimensions of objects. Neither of the two takes velocity information into account. On top of that, there is no public code or a description sufficiently detailed for reimplementation.

\section{Label Generation and Data Set} \label{section:label_generation}

As mentioned, there is no publicly available radar data set in which all types of clutter detections are annotated. For this reason, we devise a method to automatically generate the necessary labels for the public RadarScenes data set~\cite{RadarScenesDataset}. The approach is based solely on the positions and velocities of detections and the already present semantic annotations for moving objects. Additional modalities like lidar data are not needed (in contrast to \cite{Wang2021, Chamseddine2020}). The method can thus be easily applied to any other data set with similar annotations. Its code is available at \url{www.github.com/kopp-j/clutter-ds}. By providing a ready-to-run script, we effectively create a public radar clutter data set that may be adopted by other authors.

\begin{figure}[!t]
	\vspace*{0.16cm}
	\hspace*{0.825cm}
	\includegraphics[width=7.0cm]{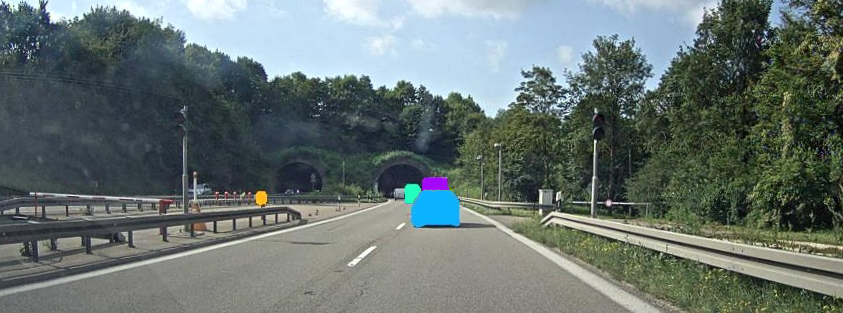}

	\tikzsetnextfilename{example_labeling}
	\begin{tikzpicture}
	[
		every axis/.append style = {label style={font=\footnotesize}, tick label style={font=\footnotesize}},
		every pin/.style = {font=\footnotesize, align=center},
		every pin edge/.style = {black, densely dashed}
	]
	\begin{axis}
	[
		width=1.0*\linewidth,
		grid=major,
		grid style={dashed,gray!30},
		xlabel=$y$ (\si{m}),
		ylabel=$x$ (\si{m}),
		x label style={yshift=0.15cm},
		x dir=reverse,
		y label style={yshift=-0.25cm},
		axis equal image,
		axis line on top,
		xmin=-70,
		xmax=70,
		ymax=100,
		ymin=0,
		colormap name=cm_radar_dl_gt
	]

		\plotRadarDetectionListGT{resources/example_data_labeling.csv}

	\end{axis}
	\end{tikzpicture}

	\vspace*{-0.1cm}
	\caption{Example of radar scans (one of each sensor) containing clutter. Points mark the positions of detections relative to the ego vehicle, arrows visualize their velocity over ground. The generated labels are indicated by color: detections annotated as \textit{moving object} are blue\includegraphics{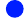}, \textit{clutter} red\includegraphics{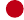} and \textit{stationary} detections gray\includegraphics{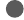}. Colored patches in the camera image cover nearby objects for legal reasons and have no meaning~\cite{RadarScenesArxiv}.}
	\label{fig:example_labeling}
\end{figure}

RadarScenes is a large automotive radar data set. It contains about \SI{4.3}{h} of manually annotated recordings. These include urban scenarios, rural routes and sequences where the ego vehicle is stationary. Four 2D \SI{77}{GHz} radar sensors are used, producing a total of \SI{119}{M} detections. Each sensor covers an area of $\pm \ang{60}$ horizontal angle and \SI{100}{m} range and faces either to the front or side of the test vehicle.
The data set's annotations focus on the positions and types of moving objects. Each detection whose position matches one of those objects is labeled with the respective class, e.g. \textit{car} or \textit{pedestrian}. All other detections are marked as background. This includes those with slight measurement errors, those stemming from stationary objects and all types of clutter.

For label generation, we define three new classes to be distinguished: \textit{moving object}, \textit{stationary} and \textit{clutter}. Radar sensors are rather inaccurate (compared, e.g., to lidar), especially regarding the azimuth angle. We thus tolerate ordinary measurement errors within reasonable bounds when deciding which detections correspond to real \textit{moving objects}. This is different from the original annotations of RadarScenes, where a detection must lie perfectly inside an object's bounding box to be assigned to the respective class. The label \textit{stationary} identifies detections stemming from nonmoving objects. The remaining detections are \textit{clutter}.

Automatic label generation works as follows. All detections originally labeled with an object class are marked as \textit{moving object}. To introduce the appropriate tolerances for measurement errors, background detections in close proximity to one of those points are searched and assigned to the class as well. This avoids the need for knowing explicit object bounding boxes. The size of each tolerance area depends on the angle of view. The maximum difference from the object detection's position is set to \SI{0.3}{m} in range and between \ang{2} and \ang{4} in azimuth, linearly increasing for higher view angles. This is because the accuracy of azimuth angles measured by the sensor degrades as the angle increases.

The remaining background detections must be split into \textit{stationary} and \textit{clutter}. But for those for which low absolute velocities were measured, it is impossible to determine if they resulted from a clutter effect, or from actual reflections at stationary objects or the ground.
However, detections with velocity below a certain threshold are usually processed by specialized methods like occupancy grid mapping \cite{Engels2021, Schumann2020, Li2018, Prophet2020}. In these, clutter is much less of an issue than e.g. for object detection. It is therefore sufficient to label only background detections above such a threshold as \textit{clutter}, and all others as \textit{stationary}.
We set the limit to $|v_\text{abs}| \ge \SI{0.5}{m/s}$. For RadarScenes, that is the velocity at which a detection is certain not to represent a stationary object. The value is composed of three times the standard deviation of the sensor's velocity measurement $3 \sigma_{v_\text{rel}} = \SI{0.3}{m/s}$ and an estimate of the error introduced during ego motion compensation.

We verify the validity of the described approach both by visually inspecting the resulting labels and experimentally, see Section~\ref{subsec:results_transfer_to_other_data}. The class distribution of the final data set is given in Tab.~\ref{tab:class_distribution}.
Exemplary sensor scans and the respective generated labels are shown in Fig.~\ref{fig:example_labeling}.

\begin{table}[!t]
	\centering
	\vspace*{0.12cm}
	\caption{Class distribution of the relabeled data set}
	\label{tab:class_distribution}
	\begin{tabular}{ c | c c c }
		\toprule
		\textbf{Class} & Moving Object & Clutter & Stationary\\
		\hline\rule{0pt}{9pt}%
		\textbf{Ratio} & \SI{3.35}{\percent} & \SI{5.57}{\percent} & \SI{91.08}{\percent}\\
		\bottomrule
	\end{tabular}
\end{table}

\section{Methods} \label{section:methods}

\subsection{Accumulation-aware Downsampling} \label{subsection:downsampling}

When employing neural networks that process point clouds directly, like PointNet++, it is beneficial to ensure a constant number of input points. This is required e.g. for batch processing, which greatly speeds up training and enables the use of batch normalization~\cite{BatchNorm}. However, sensors like lidar or radar typically produce point clouds of varying size. Fluctuations are especially pronounced for radar sensors because the number of measured detections heavily depends on the quantity and nature of objects in the environment and the sensor's velocity. For example, in the RadarScenes data set a single scan usually contains anything between $20$ and $330$ detections~\cite{SchumannThesis}. Moreover, it is common practice to increase the density of radar point clouds by accumulating scans over a sliding time window of fixed length \cite{Scheiner2021ObjectDetection}. Since consecutive scans include similar numbers of detections, this amplifies fluctuations of the point clouds' size even more. To provide the network with a set number of points regardless, resampling must be applied. When the sensors output too few detections, an upsampling scheme must artificially add data, e.g. by duplication. In the opposite case, downsampling selects only the necessary number of points. This can be done, for example, by random sampling or via a rule like removing the detections with the lowest radar cross-section (RCS) \cite{Chamseddine2020, Kraus2020}.

The problem when applying downsampling is that the network will not generate a prediction for some of the measured detections. Therefore, it is often omitted during inference, where a constant number of input points is not strictly necessary. This comes with a disadvantage for practical application, however: Outliers in the input point cloud's size are reflected in an unusually high processing time. Furthermore, the different configuration for training and inference can potentially reduce performance for such exceptionally large point clouds.

We notice that the accumulation of multiple radar scans, which is a main reason for the heavy fluctuation of point cloud sizes, also opens the possibility for new downsampling methods resolving these issues.
For use in a car or robot, the only predictions that are actually relevant are those for the most recent scan in the point cloud. Older detections have already been processed before and are included primarily to combat the sparsity of radar data.
These considerations lead to the solution of the above problems: Downsampling must be restricted to remove only detections that belong to one of the old scans in the point cloud. That way, a prediction is produced for all relevant detections even if downsampling is applied. The number of processed points can thus be kept constant also during inference.

A second, even more efficient method that achieves the same goals is the following. Instead of employing an active downsampling step, detections can be stored in a fixed-size queue, i.e. a data structure where pushing new points in at the front results in the same number of oldest points getting dropped from the back. The size of this queue should be set to the desired number of points. Then, its use is equivalent to downsampling by removing as many of the oldest scans in the accumulation time window as needed. The oldest scan remaining after that usually does not fit into memory in full. Which of its detections are kept is determined by the order in which they were initially pushed into the queue. By applying this method, the amount of memory required for buffering sensor data is reduced to a minimum and the time needed for active downsampling is saved.

Despite the simplicity of the underlying ideas, we are, to the best of our knowledge, the first to explore accumulation-aware downsampling of radar point clouds and to propose approaches like this. The only other work in which the time of measurement is even considered during downsampling is~\cite{Kraus2021}. There, discarding old detections is preferred, but some of the latest ones are likely to be removed as well.

\subsection{Network Architectures and Setup}

A basic network architecture that is well suited for processing point clouds is PointNet++~\cite{PointNet++}. Fundamentally, it recreates the encoder-decoder structure known from convolutional neural networks. In multiple so-called set abstraction (SA) levels, the number of considered points is progressively reduced by sampling. At the same time, the degree of abstraction and the area represented by each point are increased. Following this, the matching number of feature propagation (FP) levels interpolate features (and finally a classification) for all points in the original input point cloud.

The starting point of our work is the PointNet++ variant and setup described in \cite{Schumann2018, SchumannThesis}. There, the original architecture is modified for the semantic segmentation of object classes in 2D radar data. Three SA and FP levels are used, the former of which employ multi-scale grouping (MSG). The input to the network are point clouds of $3072$ detections. These are created by accumulating scans from four radar sensors over a time window of \SI{500}{ms}. All details can be found in \cite{SchumannThesis}. While it is possible to use this setup for clutter detection without major adjustments, as is done in \cite{Kraus2020} and \cite{Kraus2021}, we decide to adapt it to the task.

\begin{figure}[!t]
	\centering
	\vspace*{0.12cm}
	\includegraphics{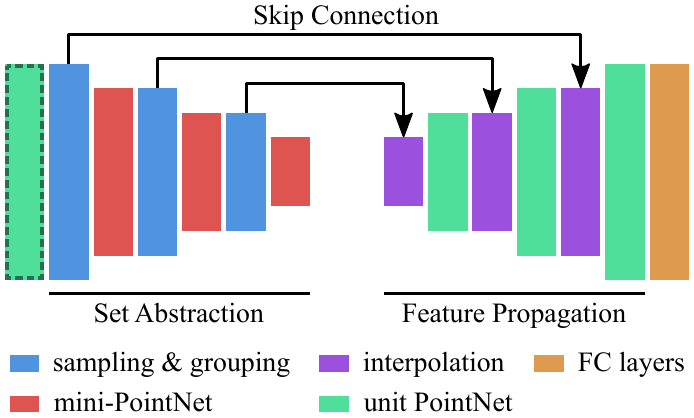}
	\caption{Fundamental network architectures employed by the two proposed variants. The preprocessing module marked by the dashed line is used only in variant B.}
	\label{fig:network_architecture}
\end{figure}

We propose two improved variants with modified input data, network parameters and training setup. The first, in the following referred to as variant A, is designed for processing point clouds of only $1280$ detections. This reduces the inference time and the required amount of memory. Variant B pushes that approach to the limit by working on single sensor scans, completely forgoing accumulation. An overview of the two network architectures is shown in Fig.~\ref{fig:network_architecture}.

Analyzing the effects that cause clutter reveals that for many of them the clutter detections and the objects involved in their formation may lie far apart (cf. Fig.~\ref{fig:overview_clutter}). For this reason, we test different ways to incorporate a wider spatial context into the calculation of features, e.g. by utilizing an attention mechanism similar to~\cite{Yan2020}. We find that increasing the radii of neighborhood regions considered during MSG works best.
For variant B, we make further modifications to compensate for the much lower density of points when processing single scans (only $144$ detections on average). In particular, we prepend the feature preprocessing module proposed in~\cite{Cennamo2020}, reduce the number of points sampled in the SA levels and add a third scale (i.e. query region) to grouping
\footnote[1]{Details: In variant A, $1024$, $512$ and $256$ points are sampled in the three SA levels, respectively. MSG uses two scales of radii $(1, 3)$, $(2, 5)$ and $(4, 10)\,\si{m}$. Layer sizes are the same as in \cite{SchumannThesis}. Variant B samples only $256$, $128$ and $64$ points. Grouping radii are $(1, 3, 6)$, $(2, 4, 8)$ and $(3, 6, 12)\,\si{m}$. The unit PointNet for preprocessing employs $(64, 64, 32)$ channels.}.

\begin{table*}[!t]
	\centering
	\vspace*{0.10cm}
	\caption{Quantitative comparison of different network setups for clutter detection. Class-specific and mean precision (Prec.), recall and F1 score (F1) are given in \si{\percent}.}
	\label{tab:results_comparison}
	\begin{tabular}{ c c | c c c | c c c | c | c | c }
		\toprule
		\multirow{2}{*}{Setup} & \multirowcell{2}{Accumulation-aware\\Downsampling} & \multicolumn{3}{c|}{Moving Object} & \multicolumn{3}{c|}{Clutter} & Stationary & \: Mean \: & \multirowcell{2}{Inference\\Time}\\
		&& Prec. & Recall & F1 & Prec. & Recall & F1 & F1 & F1 &\\
		\hline\rule{0pt}{9pt}%
		Baseline~\cite{SchumannThesis} & -- & $80.16$ & $\mathbf{94.67}$ & $86.81$ & $66.08$ & $86.47$ & $74.91$ & $98.43$ & $86.72$ & \SI{7.78}{ms}\\
		\cite{Kraus2020} & -- & $67.96$ & $92.55$ & $78.36$ & $56.91$ & $81.81$ & $67.12$ & $97.30$ & $80.93$ & \SI{6.56}{ms}\\
		\cite{Griebel2021} & -- & $83.12$ & $89.94$ & $86.39$ & $91.69$ & $88.48$ & $90.05$ & $99.55$ & $92.00$ & \textbf{\SI[detect-weight=true]{4.11}{ms}}\\
		Ours, variant A & Restriction to old points & $\mathbf{88.98}$ & $93.84$ & $\mathbf{91.34}$ & $93.11$ & $\mathbf{94.89}$ & $\mathbf{93.99}$ & $\mathbf{99.75}$ & $\mathbf{95.03}$ & \SI{7.19}{ms}\\
		Ours, variant A & Fixed-size queue & $88.36$ & $94.13$ & $91.15$ & $92.72$ & $94.76$ & $93.73$ & $99.73$ & $94.87$ & \SI{7.19}{ms}\\
		Ours, variant B & -- & $85.21$ & $91.44$ & $88.22$ & $\mathbf{93.25}$ & $92.15$ & $92.69$ & $99.74$ & $93.55$ & \SI{5.90}{ms}\\
		\bottomrule
	\end{tabular}
\end{table*}

Regarding the input to the networks, more changes beyond the reduction of point cloud sizes are made. In the original setup, each detection is described by its position, ego motion compensated velocity, RCS and timestamp. We choose to use the position in the Cartesian vehicle coordinate system (CS) at the time of recording the most recent scan in the point cloud. The timestamp is relative to that moment. Moreover, we add the detection's position in the polar sensor CS, which is more useful in some cases, and the measuring sensor's ID.
When upsampling of the input point cloud is required, that is done by duplicating randomly selected detections. Replicas created this way are excluded from loss calculation to avoid skewing gradients. For downsampling, either of the methods presented in Section~\ref{subsection:downsampling} can be used.

Finally, we also adjust the training setup. Input data is normalized via standardization. To prevent a distortion of following distance calculations, the variances of the Cartesian coordinates are averaged before scaling.
The learning rate is varied between $10^{-9}$ and $10^{-3}$ according to a cyclical learning rate policy. And we employ focal loss~\cite{FocalLoss} to increase weighting of detections that are especially hard to classify correctly and to better address the severe class imbalance. The remaining training parameters are as described in \cite{SchumannThesis}.

\section{Experiments and Results} \label{section:experiments}

We evaluate the described network variants for clutter detection regarding their performance and inference time. For this, the RadarScenes data set together with the newly generated labels is used (see Section~\ref{section:label_generation}). When a setup requires the accumulation of multiple radar scans to form a larger point cloud, only predictions for the most recent scan are analyzed, as motivated in Section~\ref{subsection:downsampling}. Points that are added by upsampling are ignored.
To minimize the influence of chance on results, we choose five different initial seeds for random number generators and repeat training of each network five times, once per seed. The mean performance of those runs on the validation set is then reported. Stated inference times are the average time a model requires for processing one point cloud on an Nvidia RTX 2080 Ti GPU.

\subsection{Comparison with Other Approaches} \label{subsec:results_comparison}

To assess the effectiveness of our setups, we compare them with other approaches in literature. As baseline, we use the setup for the semantic segmentation of objects presented in \cite{SchumannThesis}. It was the starting point for our work and is attuned specifically to RadarScenes. Except for adjustments to the input, the same network is employed for identifying multipath ghosts in \cite{Kraus2020}. We also recreate that setup and test it on our data. Finally, we evaluate the setup and best performing architecture from \cite{Griebel2021} (PointNet++ with multi-form grouping).
Regarding our own networks, variant A, which processes accumulated point clouds, is tested in combination with both proposed accumulation-aware downsampling methods. For the one that restricts downsampling to the removal of old detections, data is accumulated over \SI{300}{ms} and removed points are selected by random sampling. When instead a fixed-size queue is employed, the detections of a scan are stored in such order that the slowest are dropped first.

We follow the original training setups as closely as possible. Changes are made only when they are necessary to adapt to the new data. Accordingly, the model from \cite{SchumannThesis} is trained for $25$ epochs, others for $20$.
Class weights for loss calculation are determined the same way for all setups except \cite{Kraus2020}, which specifies its own formula to be applied. The weight of stationary detections is empirically set to $w_\text{s} = 0.6$. The other two classes are weighted proportionately to their frequencies $f_\text{o}, f_\text{c}$ for the respective configuration, i.e. such that $w_\text{o} \cdot f_\text{o} = w_\text{c} \cdot f_\text{c}$ and $\sum_i w_i \cdot f_i = 1$.

The results of the experiments are listed in Tab.~\ref{tab:results_comparison}. As can be seen, employing any of our new network variants leads to significant improvements over the existing approaches. At the same time, the reduced number of input points results in a speed-up compared to the baseline setup.
Restricting downsampling to remove only old detections performs slightly better than utilizing a fixed-size queue. However, the differences are small enough that the superior memory efficiency of the queue could still justify the tradeoff. Nonetheless, we always use the former method with variant A in the following sections.
The setups that process sensor scans individually, i.e. \cite{Griebel2021} and our variant B, work remarkably well. Despite the extreme sparsity of the point clouds and the complete lack of temporal information, they perform much better than the baseline. Our setup has the higher accuracy, and still lowers the inference time by about \SI{24}{\percent} compared to \cite{SchumannThesis}.

\begin{figure*}[!t]
	\vspace*{0.04cm}
	\begin{subfigure}{0.315\textwidth}
		\hspace*{-0.25cm}
		\tikzsetnextfilename{example_evaluation_fr_mirror}
		\begin{tikzpicture}
		[
			every axis/.append style = {label style={font=\footnotesize}, tick label style={font=\footnotesize}},
			every pin/.style = {font=\footnotesize, align=center},
			every pin edge/.style = {black, densely dashed}
		]
		\begin{axis}
		[
			width=1.26*\linewidth,
			grid=major,
			grid style={dashed,gray!30},
			xtick={-40, -20, 0, 20, 40},
			ytick={0, 20, 40, 60, 80},
			xlabel=$y$ (\si{m}),
			ylabel=$x$ (\si{m}),
			x label style={yshift=0.18cm},
			x dir=reverse,
			y label style={yshift=-0.175cm},
			axis equal image,
			axis line on top,
			xmin=-45,
			xmax=45,
			ymax=80,
			ymin=-5,
			colormap name=cm_radar_dl_confusion
		]

			\plotRadarDetectionListConfusion{resources/example_data_evaluation_fr_mirror.csv}

		\end{axis}
		\end{tikzpicture}

		\vspace*{-0.1cm}
		\caption{Drive over a four-lane bridge. Specular reflections of cars at the guardrails cause a lot of clutter. Network variant A.}
	\end{subfigure}
	\hfill
	\begin{subfigure}{0.315\textwidth}
		\hspace*{-0.35cm}
		\tikzsetnextfilename{example_evaluation_fl_chaotic}
		\begin{tikzpicture}
		[
			every axis/.append style = {label style={font=\footnotesize}, tick label style={font=\footnotesize}},
			every pin/.style = {font=\footnotesize, align=center},
			every pin edge/.style = {black, densely dashed}
		]
		\begin{axis}
		[
			width=1.26*\linewidth,
			grid=major,
			grid style={dashed,gray!30},
			xtick={-40, -20, 0, 20, 40},
			ytick={0, 20, 40, 60, 80, 100},
			xlabel=$y$ (\si{m}),
			ylabel=$x$ (\si{m}),
			x label style={yshift=0.18cm},
			x dir=reverse,
			y label style={yshift=-0.33cm},
			axis equal image,
			axis line on top,
			xmin=-60,
			xmax=60,
			ymax=100,
			ymin=-10,
			colormap name=cm_radar_dl_confusion
		]

			\plotRadarDetectionListConfusion{resources/example_data_evaluation_fl_chaotic.csv}

		\end{axis}
		\end{tikzpicture}

		\vspace*{-0.1cm}
		\caption{Highly complex scenario in urban traffic. Buildings on both sides, trees on the median strip. Network variant A.}
	\end{subfigure}
	\hfill
	\begin{subfigure}{0.315\textwidth}
		\hspace*{-0.35cm}
		\tikzsetnextfilename{example_evaluation_no_accu_fl_hard}
		\begin{tikzpicture}
		[
			every axis/.append style = {label style={font=\footnotesize}, tick label style={font=\footnotesize}},
			every pin/.style = {font=\footnotesize, align=center},
			every pin edge/.style = {black, densely dashed}
		]
		\begin{axis}
		[
			width=1.26*\linewidth,
			grid=major,
			grid style={dashed,gray!30},
			xtick={-20, 0, 20, 40, 60},
			ytick={0, 20, 40, 60, 80, 100},
			xlabel=$y$ (\si{m}),
			ylabel=$x$ (\si{m}),
			x label style={yshift=0.18cm},
			x dir=reverse,
			y label style={yshift=-0.33cm},
			axis equal image,
			axis line on top,
			xmin=-40,
			xmax=80,
			ymax=100,
			ymin=-10,
			colormap name=cm_radar_dl_confusion
		]

			\plotRadarDetectionListConfusion{resources/example_data_evaluation_no_accu_fl_hard.csv}

		\end{axis}
		\end{tikzpicture}

		\vspace*{-0.1cm}
		\caption{Left turn at an intersection with refuge island. Variant B. Even the scattered object detections on the right are identified.}
	\end{subfigure}

	\caption{Examples of point clouds and the respective network outputs. Colors indicate the ground truth and the correctness of predictions. Detections annotated as \textit{moving object}, \textit{clutter} or \textit{stationary} are drawn in light or dark blue\includegraphics{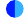}, red\includegraphics{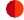} and gray\includegraphics{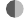}, respectively. The darker version of a color is used when the prediction matches the label, while the lighter one marks falsely classified detections. Pale green points\includegraphics{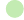} belong to old sensor scans, for which the network's output is irrelevant.
	}
	\vspace*{-0.1cm}
	\label{fig:example_evaluation}
\end{figure*}

Some exemplary data samples and the corresponding outputs of our models are visualized in Fig.~\ref{fig:example_evaluation}.

\subsection{Accumulation-aware Downsampling}

Next, we test our new accumulation-aware downsampling methods in isolation. We compare them with other ways to produce a prediction for every detection from the most recent scan in a point cloud. To this end, we repeat the evaluation of the baseline models trained in the previous section but exchange the employed method.
The original setup uses a post-processing step: Points that were removed by downsampling receive a prediction by copying it from their nearest neighbor. Another alternative is to omit downsampling entirely.

\begin{table}[!b]
	\centering
	\caption{Comparison of ways to obtain predictions for all points in the latest scan. F1 scores of baseline models are given in~\si{\percent}, inference time and its variance in \si{ms} and \si{(ms)^2}.}
	\label{tab:comparison_downsampling}
	\begin{tabular}{ c | c c | c c }
		\toprule
		\multirow{2}{*}{Method} & \multicolumn{2}{c|}{F1 Score} & \multicolumn{2}{c}{Inference Time}\\
		& Clutter & Mean & Mean & Variance\\
		\hline\rule{0pt}{9pt}%
		Post-processing \cite{SchumannThesis} & $74.91$ & $86.72$ & $7.78$ & $0.665$\\
		Omit downsampling & $90.37$ & $92.48$ & $7.79$ & $0.704$\\
		Remove only old points & $90.62$ & $\mathbf{92.50}$ & $\mathbf{7.40}$ & $\mathbf{0.582}$\\
		Fixed-size queue & $\mathbf{90.69}$ & $92.04$ & $\mathbf{7.40}$ & $\mathbf{0.582}$\\
		\bottomrule
	\end{tabular}
\end{table}

The comparison in Tab.~\ref{tab:comparison_downsampling} shows that the post-processing step from \cite{SchumannThesis} is not suited for application in clutter detection. This is because, unlike detections stemming from real objects, clutter occurs not only in local groups but also as single detections scattered between other points. Copied predictions are therefore oftentimes incorrect.
The omission of downsampling and the two accumulation-aware methods result in similar, vastly improved performance. The advantage of the latter, however, is that processed point clouds always contain the same number of detections. This reduces both the average inference time and its fluctuations. In the most extreme cases, up to about \SI{10.5}{k} detections are accumulated over the time window of \SI{500}{ms}. For this number of points, inference takes \SI{10.94}{ms} on average when downsampling is omitted. The accumulation-aware methods prevent such spikes. Furthermore, they can be applied already during training to adapt the model to the used scheme.

\subsection{Sensor-specific Models}

In the previous sections, data from all four available radar sensors is mixed. Thus, models must learn to generalize between the different mounting positions. We now investigate the effect of using a model that operates exclusively on data of a specific sensor.
Experiments are conducted for the sensor facing to the front left since the view onto oncoming traffic leads to a lot of difficult situations with multiple moving objects. For the sensor-specific version of setup variant A, we use an extended accumulation time window of \SI{1.1}{s} so that point clouds of similar size as before are obtained.

\begin{table}[!b]
	\centering
	\caption{F1 scores for data of the front left sensor (in \si{\percent}) achieved by general and sensor-specific models}
	\label{tab:results_fl_sensor}
	\begin{tabular}{ c c | c c c c }
		\toprule
		Variant & Sensor-specific & Mov. Obj. & Clutter & Statio. & Mean\\
		\hline\rule{0pt}{9pt}%
		A && $\mathbf{92.50}$ & $\mathbf{92.58}$ & $99.70$ & $\mathbf{94.93}$\\
		A & $\checkmark$ & $91.81$ & $91.68$ & $\mathbf{99.71}$ & $94.40$\\
		B && $89.63$ & $90.50$ & $99.67$ & $93.27$\\
		B & $\checkmark$ & $89.90$ & $90.40$ & $99.66$ & $93.32$\\
		\bottomrule
	\end{tabular}
\end{table}

The performance for data of the front left sensor of general and sensor-specific versions of our setups is reported in Tab.~\ref{tab:results_fl_sensor}.
Contrary to what one might expect, performance of variant B is barely increased by training specifically for the selected sensor. Using the sensor-specific version of variant A even leads to a loss of accuracy. This suggests that a short accumulation of data from multiple sensors with overlapping fields of view is preferable to treating sensors separately.

\subsection{Transfer of Models To Other Data and Comparison with Rule-based Approach} \label{subsec:results_transfer_to_other_data}

In an earlier work, we recorded a small radar clutter data set with our own test vehicle.
Details can be found in \cite{Kopp2021}. Since here the correctness of all annotations was manually checked, this data can be utilized to validate the proposed label generation method.

\begin{table}[!b]
	\centering
	\caption{Comparison of approaches on the data set from~\cite{Kopp2021}. Models are pretrained on RadarScenes and then fine-tuned on the new data. Reported values are F1 scores measured using 4-fold cross-validation in \si{\percent}.}
	\label{tab:results_finetuning_radar_clutter_ds}
	\begin{tabular}{ c | c c c c }
		\toprule
		Approach & \makecell{Nonclutter \&\\Ambiguous} & Clutter & Stationary & Mean\\
		\hline\rule{0pt}{9pt}%
		Ours, variant A & $\mathbf{85.66}$ & $\mathbf{96.18}$ & $\mathbf{99.63}$ & $\mathbf{93.82}$\\
		Ours, variant B & $80.51$ & $94.00$ & $99.40$ & $91.30$\\
		Rule-based~\cite{Kopp2021} & $44.36$ & $87.78$ & -- & $66.07$\\
		\bottomrule
	\end{tabular}
\end{table}

The best performing model of setup A from \mbox{Section~\ref{subsec:results_comparison}}, which was trained only on the relabeled RadarScenes data set, achieves a mean F1 score of \SI{93.72}{\percent} for the new data. This is especially remarkable considering that just a single sensor with a different mounting position and orientation than before is used. The only adjustment to the network setup that is needed is an extension of the accumulation time to \SI{500}{ms}.
Interestingly, performance can hardly be further improved by fine-tuning the model on the new data, cf. Tab.~\ref{tab:results_finetuning_radar_clutter_ds}. Together with the impressive results right from the start, this shows that the method we devised for label generation is highly accurate. A model trained with the produced annotations learns to identify the characteristics of classes sufficiently well that generalization to data of a different sensor configuration is easily possible.

The evaluation on the second data set has an additional benefit. It enables a direct comparison of our new setups with the rule-based approach described in \cite{Kopp2021}, see again Tab.~\ref{tab:results_finetuning_radar_clutter_ds}. Even though that algorithm has to make a prediction only for points with significant velocity, the neural networks exhibit much better performance.
Their drawback, of course, is their comparatively high computational cost and innate lack of explainability.

\section{Conclusion} \label{section:conclusion}

In this work, we present a method to automatically generate ground-truth for a radar clutter data set. It requires only the basic sensor outputs and the object annotations of freely available data. The method's code is published. This results in the first data set of realistic driving scenarios in which clutter is labeled that is accessible to all researchers.
We then modify an existing setup of PointNet++ for the detection of clutter in automotive radar point clouds. The number and form of input points, the architecture and the training setup are all tailored specifically to the task. By introducing accumulation-aware downsampling, we further increase efficiency.
Our best performing setup detects almost \SI{95}{\percent} of clutter while only rarely misclassifying other points.
This is a significant improvement over existing machine-learning approaches and a comparable rule-based method. An alternative setup variant reduces the inference time considerably and still manages to achieve similar performance.

\bibliographystyle{IEEEtran}
\bibliography{references}

\end{document}